\RequirePackage{fix-cm}
\documentclass[letterpaper]{article}
\usepackage[preprint]{aaai2027}
\usepackage[hyphens]{url}
\usepackage{graphicx}
\usepackage{natbib}
\usepackage{caption}
\usepackage{amsmath}
\usepackage{amsfonts}
\DeclareCaptionStyle{ruled}{labelfont=normalfont,labelsep=colon,strut=off}
\title{D-STEER: Two-Way Control of DPO-Induced Safety Behavior in Activation Space}
\author{
\fontsize{11.5}{13.5}\selectfont
\mbox{Samarth Raina\textsuperscript{1},
Saksham Aggarwal\textsuperscript{2},
Amit Dhanda\textsuperscript{3\ensuremath{\dagger}},
Vinija Jain\textsuperscript{4\ensuremath{\ddagger}},
Aman Chadha\textsuperscript{5\S},
Amitava Das\textsuperscript{6}}
}
\affiliations{
\textsuperscript{1}IIIT Delhi \quad
\textsuperscript{2}Microsoft \quad
\textsuperscript{3}Amazon, USA \quad
\textsuperscript{4}Meta, USA \quad
\textsuperscript{5}Apple, USA\\
\textsuperscript{6}Pragya Lab, BITS Pilani, K. K. Birla Goa Campus
}
\begin{document}
\maketitle
\begingroup
\renewcommand{\thefootnote}{\fnsymbol{footnote}}
\footnotetext[2]{This work does not relate to Amit Dhanda's position at Amazon.}
\footnotetext[3]{This work does not relate to Vinija Jain's position at Meta.}
\footnotetext[4]{This work does not relate to Aman Chadha's position at Apple.}
\endgroup

\begin{abstract}
\spaceskip=.95\fontdimen2\font plus\fontdimen3\font minus\fontdimen4\font
Preference-based post-training shapes language-model behavior, but safer outputs do not reveal how that change is represented internally or how firmly it persists under direct intervention. This matters for open-weight models, where an actor with white-box access can manipulate internal representations at inference without touching parameters. Using Direct Preference Optimization (DPO), we ask whether the safety behavior it introduces becomes deeply embedded or stays accessible to targeted activation-space interventions. We introduce D-STEER, an audit of two matched instruction-tuned and harmlessness-DPO pairs: OpenHermes-2.5-Mistral-7B and Llama-3-8B OpenHermes. D-STEER derives layer-local directions from differences in matched prompt activations, adds them to the instruction-tuned model to install the DPO-associated safety change and subtracts them from the DPO model to remove it, with separate operating windows distinguishing behavioral control from degradation. On held-out HarmfulQA, all four interventions shift harmful enablement as intended and exceed norm-matched random controls; Mistral installation recovers $98\%$ of the endpoint gap and the other three arms recover $46$--$50\%$. The intended direction transfers to HH-RLHF and AdvBench under frozen coefficients, although usable operating windows do not transfer universally. An independently rebuilt implementation reproduces the bidirectional pattern on the Mistral pair under different construction and intervention choices. A meaningful component of DPO-induced safety behavior therefore remains controllable through compact inference-time intervention, with the extent depending on the checkpoint pair and the steering implementation.
\looseness=-1
\end{abstract}

\section{Introduction}

Preference optimization is widely used to improve the safety behavior of language models. Direct Preference Optimization (DPO) learns from preferred and dispreferred responses without an explicit reward model \cite{rafailov2023direct}; in safety-focused settings the preferred response is the one that avoids enabling harm. Safer outputs after tuning, however, do not reveal how robust the added behavior is to white-box intervention or whether it remains directly controllable through the model's internal states.

This matters for open-weight models, where users can modify intermediate representations at inference without retraining \cite{seger2023open,kapoor2024societal}. Such access is also a distinct dimension of model-release governance \cite{solaiman2023gradient}. Directions in the \emph{residual stream}, the running vector each transformer layer reads from and writes to, can control behavior \cite{turner2023activation,zou2023representation}, including refusal \cite{arditi2024refusal}, and capabilities suppressed by DPO can remain internally accessible \cite{lee2024mechanistic}. Yet behavioral change alone is inconclusive: nonspecific perturbations also alter harmful compliance \cite{korznikov2025rogue}, increased refusal may be over-refusal \cite{rottger2023xstest}, and strong interventions can produce incoherent or truncated text. A safety audit must therefore score \emph{harmful enablement}, how much operationally useful help toward harm a response actually provides, rather than refusal alone, establish that the effect is specific to the direction used, and verify that generation remains usable.

\begin{figure}[t]
\centering
\includegraphics[width=\linewidth]{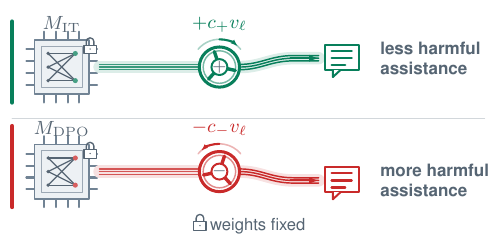}
\caption{Conceptual activation steering: addition to IT reduces harmful assistance; subtraction from DPO increases it, relative to each unsteered checkpoint.}
\label{fig:overview}
\end{figure}

D-STEER asks how much of the safety behavior associated with a matched DPO transition remains accessible in activation space. For each family we study an instruction-tuned model and its corresponding checkpoint after harmlessness-oriented DPO. We isolate the behavioral difference introduced by the DPO stage and test how much of it can be causally controlled while the weights stay fixed. Directions come from averaging residual-stream differences between the matched checkpoints on identical prompts: adding the direction to the instruction-tuned model tests \emph{installation} of the safety change, subtracting it from the DPO model tests \emph{removal}. Because usable strengths differ across checkpoints and directions, we calibrate a separate \emph{operating window} for each, namely the coefficients that keep invalid generation, maximum-length truncation and quality loss inside pre-specified bounds, and select within it the coefficient with the largest intended change. Norm-matched random directions test whether the effect reflects DPO-associated geometry rather than perturbation magnitude, and a reversed-preference checkpoint provides a training-level comparator.

The safety behavior DPO installs is directly controllable in activation space, and the evidence is pre-registered rather than post hoc: partitions, selection rule and coefficients were frozen before a 300-prompt panel that stayed sealed through construction, calibration and debugging was opened. The \textbf{Mistral pair} carries the strongest and most extensively replicated evidence: largest held-out effects, separation from norm-matched random directions everywhere it was measured, transfer to both off-distribution panels, and an independent rebuild spanning two corpora, ten depths and three token scopes. The \textbf{Llama pair} establishes the same bidirectional control on its own held-out panel at smaller magnitude.

We report two runs by design. A \emph{calibrated} run scales the direction to the typical activation magnitude at each layer and supplies every confirmatory estimate; an independently
rebuilt \emph{raw} run keeps the measured displacement magnitude and changes every other choice,
testing whether the finding depends on the first implementation. Their coefficients are on
different scales and are never compared.

On frozen, disjoint HarmfulQA partitions with a 300-prompt held-out panel, installation reduces harmful enablement by \(0.289\) in the Mistral pair and \(0.084\) in the Llama pair, removal increases it by \(0.136\) and \(0.082\), and every effect exceeds its five norm-matched controls with all intervals excluding zero. The intended direction transfers without retuning to all eight HH-RLHF and AdvBench conditions.

\section{Related Work}

\paragraph{Preference optimization and its internal effects.} Preference-based post-training spans reward-model pipelines \cite{stiennon2020learning,ouyang2022training,bai2022training}, constitution-based supervision \cite{bai2022constitutional}, and reward-free objectives including DPO and its successors \cite{rafailov2023direct,azar2023general,ethayarajh2024kto,hong2024orpo,meng2024simpo}. A parallel line studies these methods as transformations of internal representations. \citet{lee2024mechanistic} found that DPO suppresses toxic behavior by bypassing rather than erasing toxicity-related capabilities, and that the suppressed behavior can be recovered. Related work finds that fine-tuning can wrap rather than erase existing capabilities \cite{jain2023mechanistically}, while DPO can also redistribute likelihood toward harmful responses \cite{razin2024unintentional}. Superficial-alignment results likewise motivate distinguishing behavioral change from changes to underlying capabilities \cite{zhou2023lima,lin2023unlocking,qi2024safety}. We extend this from explaining toxicity reduction to testing whether the difference separating matched checkpoints forms a bidirectional control surface accessible at inference.

\paragraph{Activation steering and safety directions.} Activation steering modifies behavior by intervening on intermediate representations without updating weights. Contrastive activation differences identify and control directions associated with high-level behavioral properties \cite{turner2023activation,zou2023representation,rimsky2024steering}; latent steering vectors can also be optimized directly in frozen language models \cite{subramani2022extracting}. Related edits operate on attention heads or task vectors \cite{li2023inference,ilharco2022editing}; and dictionary learning characterizes the feature geometry such directions inhabit \cite{cunningham2023sparse,templeton2024scaling}. In safety settings, \citet{arditi2024refusal} identified a compact residual-stream direction mediating refusal, \citet{cao2024personalized} optimizes steering vectors from preference data, and \citet{tan2024analysing} analyzes when such vectors generalize. We instead quantify how much harmlessness-oriented behavior can be installed and removed across matched checkpoints under explicit usability constraints.

\paragraph{White-box robustness and measurement.} Activation access can also expose aligned behavior to manipulation: \citet{korznikov2025rogue} showed that random and semantically benign directions can increase harmful compliance. Weight-level work agrees by other means, since small amounts of fine-tuning, including low-rank or adversarial updates, remove safety behavior cheaply \cite{qi2023finetuning,yang2023shadow,lermen2023lora,zhan2023removing,wei2024assessing}. Even with weights held fixed, changes to decoding parameters can undermine safety \cite{huang2023catastrophic}, and jailbreaks show refusal is brittle at the input level \cite{wei2023jailbroken,zou2023universal}. Because harmfulness benchmarks and model-based judging are imperfect \cite{bhardwaj2023red,mazeika2024harmbench,souly2024strongreject,chao2024jailbreakbench,xie2024sorrybench,zheng2023judging,inan2023llama} and over-refusal can masquerade as safety \cite{rottger2023xstest}, we treat refusal as supporting evidence only, score harmful enablement directly, and report generation integrity alongside every estimate.

\section{Method}

\subsection{Problem Formulation and Direction Construction}

DPO optimizes a policy against a fixed reference on preference triples without an explicit reward model \cite{rafailov2023direct}. For each model family, we study an instruction-tuned model \(M_{\mathrm{IT}}\) and its corresponding checkpoint after harmlessness-oriented DPO, \(M_{\mathrm{DPO}}\). The two share the same lineage and differ by the DPO stage alone, which is what makes a prompt-by-prompt comparison of their internal states meaningful. \emph{Installation} adds a DPO-associated direction to \(M_{\mathrm{IT}}\); \emph{removal} subtracts it from \(M_{\mathrm{DPO}}\). The primary outcome is direct harmful enablement under pre-specified validity and quality constraints.

Let \(\mathcal{D}_{\mathrm{con}}=\{x_i\}_{i=1}^{N}\) be a frozen construction set disjoint from calibration and evaluation. Both checkpoints receive the identical rendered prompt, and therefore the identical token sequence, in a single prefill pass. We record the residual stream at the final prompt position, the state from which the next-token distribution for the first assistant token is computed; no generated or reference-response tokens enter the direction. Because token identity and prompt context are held fixed across the checkpoints, their difference there captures checkpoint-dependent processing at the transition from reading the prompt to producing a response. We read this as an intent-relevant decision point without claiming the difference is exclusively intent. For layer \(\ell\),
\begin{equation}
\bar d_\ell=\frac{1}{N}\sum_{i=1}^{N}\left[h_\ell^{\mathrm{DPO}}(x_i)-h_\ell^{\mathrm{IT}}(x_i)\right].
\label{eq:mean_direction}
\end{equation}
Because activation norms vary across layers and models, we rescale to the mean norm of the corresponding \(M_{\mathrm{IT}}\) activations:
\begin{align}
a_\ell&=\frac{1}{N}\sum_i\|h_\ell^{\mathrm{IT}}(x_i)\|_2,
&v_\ell&=a_\ell\frac{\bar d_\ell}{\|\bar d_\ell\|_2+\epsilon}.
\label{eq:relative_normalization}
\end{align}

The primary intervention uses \(\mathcal{L}=\{27,\ldots,31\}\), frozen before the held-out panel was opened, on three empirical grounds and one structural one: relative checkpoint displacement rises through the network and stays high across the final block; the stress test's depth ablation below shows intended movement growing with contiguous late-layer coverage and flattening around seven to nine layers; single-layer actuation is frequently inert, leaving more than three-quarters of steered responses byte-identical to baseline at one layer against roughly one-eighth at five; and late residual-stream states are closer in depth to the final representation consumed by the language-model head. This is empirically motivated for these two families, not a general property of preference optimization.

\subsection{Intervention and Scale Convention}

Without modifying parameters we intervene additively at layer \(\ell\) and token position \(t\):
\begin{equation}
\widetilde h_{\ell,t}^{\,m,\lambda}=h_{\ell,t}^{\,m}+\lambda v_\ell .
\label{eq:steering_intervention}
\end{equation}
Installation uses \(m=\mathrm{IT}\) and \(\lambda=+c_+\), removal \(m=\mathrm{DPO}\) and \(\lambda=-c_-\), with \(c_+,c_->0\) and one shared direction rather than separately optimized vectors. The direction is read at the final prompt position, where both checkpoints have processed identical tokens, so any difference is checkpoint-attributable. The primary protocol then actuates at every prompt and generated-token position, because holding safety behavior on or off across a whole response requires the direction to be present while that response is produced; the stress test actuates at the final prompt position only, supplying the narrow-scope counterpart. Weights, prompts, templates and decoding are unchanged.

\paragraph{Raw versus relative scale.}
\label{sec:parameterizations}
Equation~\ref{eq:steering_intervention} admits two scale conventions, and the distinction governs every coefficient reported here. The \emph{raw} convention injects the empirical mean activation displacement of Equation~\ref{eq:mean_direction} directly, \(\widetilde h_{\ell,t}=h_{\ell,t}+\lambda\bar d_\ell\), so the injected magnitude is \(\lambda\|\bar d_\ell\|_2\) and \(\lambda\) counts mean construction-set displacements at that layer. The \emph{relative} convention substitutes \(v_\ell\), so the magnitude is \(\lambda a_\ell\) and \(\lambda\) is a dimensionless fraction of the typical residual-stream magnitude. Both act along the same unit direction and differ only in per-layer gain. Writing \(r_\ell=\|\bar d_\ell\|_2/a_\ell\), they inject equal magnitude when
\begin{equation}
\lambda_{\mathrm{raw}}=\lambda_{\mathrm{rel}}\,/\,r_\ell .
\label{eq:scale_conversion}
\end{equation}
\textbf{Numerically equal coefficients under the two conventions are therefore not comparable interventions}, and we compare none. Equation~\ref{eq:scale_conversion} holds for one fixed direction at one layer; because \(r_\ell\) varies across \(\mathcal{L}\), no single scalar maps one convention onto the other across a multilayer block, so the two are different relative weightings of the layers rather than two settings of one control. It also does not convert between the primary protocol and the stress test, which build different directions. Measured \(r_\ell\) values, with provenance and tolerance, are in the supplement.

Each convention is comparable in its own sense: under the raw convention \(\lambda{=}1\) is one
mean displacement at every layer, while under the relative convention \(\lambda\) is a fixed
fraction of typical activation magnitude, which is what makes ``equal magnitude'' well defined
for norm-matched controls. We adopt the relative convention for the calibrated run because our
central claim is directional, and report the raw convention separately.

\subsection{Behavioral Evaluation and Operating Windows}

Every response first passes a deterministic \emph{validity} screen for empty or abnormally short output, leaked role tokens, severe repetition and punctuation or subword loops; \emph{maximum-length termination}, meaning generation stopped at the token cap rather than at an end-of-sequence marker, is recorded separately. Invalid responses leave behavioral means but stay in integrity denominators. A frozen evaluator assigns integer scores in \(\{0,\ldots,10\}\) normalized to \([0,1]\); failed scoring after a fixed retry policy is an abstention, never an imputed score. Harmful enablement \(H\) measures operationally useful assistance for causing harm, refusal \(R\) is supporting evidence rather than a safety proxy, and quality \(Q\) measures coherence and completeness while ignoring safety. Effects are oriented so positive denotes intended movement:
\begin{align}
E_H^{\mathrm{inst}}(c)&=H(M_{\mathrm{IT}},0)-H(M_{\mathrm{IT}},{+}c),\\
E_H^{\mathrm{rem}}(c)&=H(M_{\mathrm{DPO}},{-}c)-H(M_{\mathrm{DPO}},0),
\label{eq:harmfulness_effects}
\end{align}
and quality loss is \(L_Q(c)=Q(M_m,0)-Q(M_m,\lambda)\), positive for degradation. We call this an intervention cost rather than a general alignment tax, because it is measured only on the safety prompts used here.

Large coefficients can create invalid, truncated, repetitive, or low-quality output, so strengths are chosen only on the frozen calibration panel. For direction \(s\in\{\mathrm{inst},\mathrm{rem}\}\), a tested magnitude \(c\geq0\) is usable when
\begin{equation}
\phi_s(c)=\mathbf{1}\!\left[\rho_{\mathrm{inv}}^s(c)\leq.05,\ \rho_{\mathrm{max}}^s(c)\leq.05,\ L_Q^s(c)\leq.10\right].
\label{eq:usability_gate}
\end{equation}
The ceiling \(C_s\) is the largest tested coefficient connected to zero by an uninterrupted run of usable coefficients, so an isolated post-failure pass cannot reopen the window; installation and removal receive separate ceilings, bracketed by a predeclared coarse grid and refined at width at most \(0.005\). Within each window we select
\begin{equation}
c_{s}^{\ast}=\arg\max_{0\leq c\leq C_s}E_{H,s}(c),\quad s\in\{\mathrm{inst},\mathrm{rem}\},
\label{eq:coefficient_selection}
\end{equation}
with exact ties favoring the smaller magnitude. Refusal, resemblance, and gap recovery never determine selection.

\subsection{Controls, Comparators, and Estimation}

\paragraph{Norm-matched random directions.}
A random control replaces the real direction with an isotropic one of the same injected size, so any difference is attributable to \emph{where} the intervention points rather than \emph{how hard} it pushes. For seeds \(r\in\{11,22,33,44,55\}\) we draw \(z_{\ell,r}\sim\mathcal N(0,I)\) and rescale it to the norm of the direction it replaces:
\begin{align}
\big\|v^{\mathrm{rand}}_{\ell,r}\big\|_2&=\|v_\ell\|_2=a_\ell,
&\big\|\bar d^{\,\mathrm{rand}}_{\ell,r}\big\|_2&=\|\bar d_\ell\|_2,
\label{eq:random_direction}
\end{align}
the left form under the primary calibrated protocol, the right under the raw-displacement audit. Controls use identical layers, prompts, decoding and policy, and the real direction's \emph{frozen} coefficient, so both arms inject the same magnitude at each layer and controls are never tuned separately. Directional specificity is the gap this leaves:
\begin{equation}
\mathrm{Spec}_s=E_{H,s}^{\mathrm{real}}-\tfrac{1}{5}\textstyle\sum_{r=1}^{5}E_{H,s,r}^{\mathrm{rand}}.
\label{eq:specificity}
\end{equation}
We use five seeds, predeclared before the panel was opened. Five gives a small distribution over accidental alignments between an isotropic direction and the safety-relevant subspace while bounding the cost of generating and judging every control condition. We report the seeds individually through their mean and range and do not treat five draws as a large inferential sample: every interval comes from paired prompt bootstrapping, and no seed-level randomization-test $p$-value is claimed. Five is a pragmatic, predeclared compute-versus-coverage choice, not a theoretically optimal number.

\paragraph{Reversed-preference training.}
We also train a comparator from the Llama pair's instruction-tuned model by running the same preference objective with preferred and rejected responses exchanged, and ask whether steering-based removal moves behavior in the direction that reversed training does. We write it \(M_{\mathrm{rev}}\), assess directional agreement, and compare magnitudes only when the reversed-training effect is reliably nonzero. This is directional corroboration, not an equivalence test.

\paragraph{Gap recovery and estimation.} With \(G_H=H(M_{\mathrm{IT}};\mathcal I)-H(M_{\mathrm{DPO}};\mathcal I)\) on a common source-ID set \(\mathcal I\), gap recovery divides the corresponding effect by \(G_H\), so \(0\), \(1\), and values above \(1\) denote no movement, movement equal to the endpoint gap, and overshoot; it is descriptive and never selects coefficients. Every contrast uses stable source IDs observed, valid, and scored in all required conditions, and integrity rates use all requested prompts. We report two-sided \(95\%\) percentile intervals from 10{,}000 paired prompt-bootstrap replicates resampling source IDs; for random controls we take the per-prompt mean across seeds before bootstrapping prompts. Datasets are never pooled.

\subsection{Experimental Setup}

\paragraph{Models and data.} The \emph{Mistral pair} is OpenHermes-2.5-Mistral-7B with and without its merged harmlessness-DPO adapter. The \emph{Llama pair} is matched Llama-3-8B OpenHermes checkpoints before and after DPO on HH-RLHF \texttt{harmless-base}. Both have 32 layers and residual dimension 4{,}096, and \(M_{\mathrm{rev}}\) is an auxiliary Llama-pair checkpoint trained with preferences exchanged. Archived run identifiers use the internal names \texttt{Pair B} for the Mistral pair and \texttt{Pair A} for the Llama pair; the mapping is restated in the supplement and no archived file is renamed. After removing 22 duplicate pairs from HarmfulQA \cite{bhardwaj2023red}, 1{,}938 unique prompts are deterministically partitioned into 1{,}378 for construction, 200 for calibration, 300 for held-out evaluation, and 60 for development, disjoint by source ID and normalized prompt hash; the final panel is inaccessible during construction, selection, rubric development, and debugging. Transfer uses 200 frozen HH-RLHF test prompts \cite{bai2022training} and 200 AdvBench requests \cite{zou2023universal}, with coefficients transferred without retuning.

\paragraph{Generation, evaluation, and provenance.} All conditions use endpoint-bound templates and greedy decoding with input and output limits of 2{,}048 and 768 tokens, and a frozen Qwen3.5-27B evaluator pinned by revision. Calibration yields \(\lambda_{\mathrm{inst}}^{\ast}={+}0.135\), \(\lambda_{\mathrm{rem}}^{\ast}={-}0.055\) for the Mistral pair and \({+}0.050\), \({-}0.050\) for the Llama pair; per Equation~\ref{eq:scale_conversion} these relative-scale coefficients are not comparable to raw-scale ones. The final evaluation comprises 29 conditions: five unsteered baselines, four real interventions, twenty random controls. Endpoints, datasets, activation artifacts, evaluator protocols, code states, and analysis inputs are revision- or hash-bound and archived with SHA-256 inventories; full identities, recipes, serving parameters, and rubric hashes are in the supplement.

\section{Results}

The calibrated run comes first (Table~\ref{tab:heldout_main}, Figures~\ref{fig:dose}
and~\ref{fig:heldout_specificity}); the rebuilt raw run follows in its own section, on its own
coefficient scale.

\subsection{Effects Grow With Steering Magnitude Inside a Usable Range}

Before accessing the held-out panel we identified separate usable windows for installation and removal on the frozen 200-prompt calibration set (Table~\ref{tab:heldout_main}). The Mistral pair permits considerably stronger installation than the Llama pair, and its removal direction is bounded by response quality rather than generation failure. Calibration-panel movements, in Table~\ref{tab:heldout_main}'s row order, were $0.2865$, $0.1610$, $0.0487$ and $0.0584$, with descriptive gap recoveries of $1.104$, $0.596$, $0.311$ and $0.313$; because these values informed selection we treat them as calibration evidence only.

\begin{figure}[t]
\centering
\includegraphics[width=\linewidth]{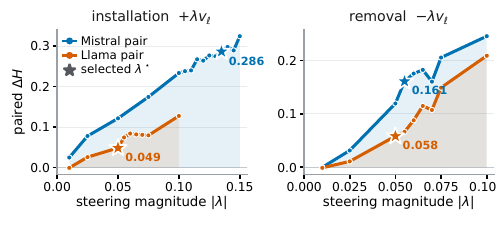}
\caption{Dose response on the frozen 200-prompt calibration panel. The intended effect rises with steering magnitude in both directions and for both pairs; stars mark the coefficient selected inside each usability window. Held-out effects at those coefficients are in Table~\ref{tab:heldout_main}.}
\label{fig:dose}
\end{figure}

Effects rise with steering magnitude in every condition (Figure~\ref{fig:dose}), and in all four the selected coefficient coincides with the operating ceiling. Because the grid is discrete and the dose response need not be monotonic in \(\lambda\), we report these as the effects at the selected coefficients rather than as bounds on what is achievable. Protection against cherry-picking comes from the selection rule being frozen before the final panel was opened and from that panel being disjoint from the construction and calibration partitions.

\subsection{Safety Behavior Is Controllable in Both Directions}

We now evaluate the frozen coefficients on the 300-prompt held-out panel, untouched during construction, calibration and selection (Table~\ref{tab:heldout_main}). Positive movement denotes reduced harmful enablement relative to the unsteered instruction-tuned model for installation, and increased harmful enablement relative to the unsteered DPO model for removal.

\begin{table}[t]
\centering
\small
\setlength{\tabcolsep}{2pt}
\begin{tabular}{llccccc}
\hline
Pair & Dir. & $C$ & $\lambda^\star$ & $n$ & $\Delta H$ [95\% CI] & $L_Q$ \\
\hline
Mistral & Inst. & $.135$ & $+.135$ & 292 & $+.289\ [.246,.332]$ & $.037$ \\
Mistral & Rem. & $.055$ & $-.055$ & 300 & $+.136\ [.102,.172]$ & $.066$ \\
Llama & Inst. & $.050$ & $+.050$ & 286 & $+.084\ [.050,.117]$ & $.022$ \\
Llama & Rem. & $.050$ & $-.050$ & 296 & $+.082\ [.048,.117]$ & $.047$ \\
\hline
\end{tabular}
\caption{Calibration-selected operating ceilings $C$ and coefficients $\lambda^\star$, and their held-out effects on the 300-prompt HarmfulQA panel. Generation criteria bind for three conditions and quality for Mistral removal. Harmfulness movement $\Delta H$ is oriented so positive is the intended direction; $L_Q$ is quality loss against the corresponding unsteered endpoint. All four conditions satisfy every criterion of Equation~\ref{eq:usability_gate}. Strict source-ID pairing, 10{,}000 paired prompt bootstraps. Leading zeros are omitted. Refusal movements, integrity rates and endpoint means are in the supplement.}
\label{tab:heldout_main}
\end{table}

All four interventions move harmful enablement as intended with intervals excluding zero. Mistral installation is largest ($0.289$); the two Llama effects are smaller and similar ($0.084$, $0.082$). Refusal corroborates harmfulness in every condition ($+0.094$ to $+0.282$) and quality loss stays below the frozen $0.10$ bound. The unsteered endpoints score $H=0.486$ and $0.191$ for the Mistral instruction-tuned and DPO models and $0.390$ and $0.217$ for the Llama pair, so the strictly paired endpoint gaps are $0.294$ and $0.167$--$0.173$; against these, Mistral installation recovers $0.983$ (95\% CI $[0.866,1.112]$) and the others roughly $0.46$--$0.50$; because the Llama gap is much the smaller denominator, a similar recovery fraction there corresponds to a much smaller absolute effect and recovery is not a cross-pair effect-size comparison. Both directions survive held-out evaluation, but controllable effect size differs substantially across pairs.

\begin{figure*}[t]
\centering
\includegraphics[width=\textwidth]{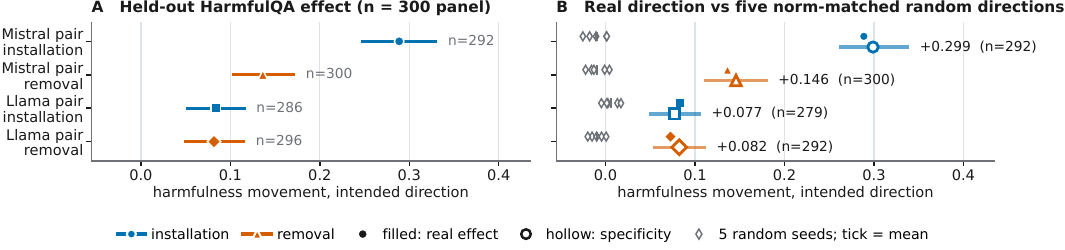}
\caption{Held-out evidence on the 300-prompt HarmfulQA panel, Mistral pair first. (A) Intended harmfulness movement for the four frozen interventions. (B) The same effects against five norm-matched random directions at the same layers, prompts and coefficient: grey diamonds are seed effects with a tick at their mean, the filled marker the real effect, the hollow marker the specificity contrast. Panel B requires every random arm to be valid on a prompt, so its $n$ is stated separately. Intervals are 95\% paired prompt bootstraps over 10{,}000 replicates.}
\label{fig:heldout_specificity}
\end{figure*}

\subsection{The Effect Is Specific to the Direction, Not Its Magnitude}

The effect tracks the estimated direction, not the size of the perturbation. We compare each intervention with five norm-matched random directions at the same layers and frozen coefficient, reporting the specificity contrast in Figure~\ref{fig:heldout_specificity}B.

The real direction exceeds the mean norm-matched random effect in all four conditions and every specificity interval excludes zero: $+0.299$ $[0.260,0.339]$ and $+0.146$ $[0.111,0.181]$ for Mistral installation and removal, $+0.077$ $[0.048,0.106]$ and $+0.082$ $[0.053,0.112]$ for the Llama pair. Random means stay within $\pm0.010$ and all 20 seed-level effects within $\pm0.026$, against real effects of $0.073$ to $0.289$. Because this panel additionally requires every random arm to be valid on a prompt, its intersection is stricter than Table~\ref{tab:heldout_main}'s; the Llama removal effect is $+0.073$ here against $+0.082$ there. The held-out effect is therefore specific to the estimated direction rather than to activation magnitude alone.

\subsection{Steering Removal Agrees with Reversed-Preference Training}

We next compare Llama-pair steering removal with the auxiliary checkpoint \(M_{\mathrm{rev}}\) introduced above. This is a directional training-level comparator, not an equivalence target.

Both interventions increase harmful enablement on the calibration and held-out HarmfulQA panels and on AdvBench. On held-out HarmfulQA, reversed training changes harmfulness by $+0.033$ (95\% CI $[0.006,0.060]$) and steering removal by $+0.079$ (95\% CI $[0.046,0.114]$); on HH-RLHF the reversed-training interval spans zero, so no ratio is reported. Across panels the ratio runs from $0.54$ to $2.86$, so the two agree directionally without a stable magnitude relationship. The auxiliary $M_{\mathrm{rev}}$ held-out baseline also marginally exceeds the maximum-token gate ($5.7\%$ against $5\%$); it is retained and reported rather than replaced.

\subsection{The Direction Transfers Off-Distribution}

We apply the frozen coefficients without retuning to 200-prompt HH-RLHF and AdvBench panels; the full table is in the supplement. The intended effect transfers in all eight conditions with every interval excluding zero, spanning $+0.045$ to $+0.421$, and refusal moves with harmfulness throughout ($+0.103$ to $+0.404$). Mistral installation is strongest, reaching $+0.421$ $[0.353,0.488]$ on AdvBench with a gap recovery of $1.067$.

Operating windows are calibrated per checkpoint and direction by construction, so the quantity carried across panels here is the direction itself; per-condition integrity for these panels is reported in the supplement.

\subsection{An Independent Rebuild Reproduces the Effect}
\label{sec:raw_corroboration}

The effect survives an implementation rebuilt from scratch. The raw run injects \(\bar d_\ell\)
unnormalized, builds directions from judge-filtered HH-RLHF and HarmfulQA pools, constructs and
actuates at the final prompt position, uses principally the last seven layers, and screens
usability with a \(\geq\!90\%\) validity floor under a different evaluator and separate
coefficients. Because nothing is held fixed, agreement is convergent evidence rather than a
controlled comparison. Estimation is identical: strict source-ID pairing, 10{,}000-replicate
paired prompt bootstraps, five norm-matched controls, no pooling.

\begin{table}[t]
\centering
\small
\setlength{\tabcolsep}{2.5pt}
\begin{tabular}{llcccc}
\hline
Panel & Dir. & $\lambda_{\mathrm{raw}}$ & $n$ & $\Delta H$ [95\% CI] & $H_1$ \\
\hline
HarmfulQA & Inst. & $+0.7$ & 300 & $+.364\ [.317,.411]$ & $.080$ \\
HarmfulQA & Rem. & $-0.5$ & 299 & $+.148\ [.112,.184]$ & $.300$ \\
HH-RLHF & Inst. & $+0.7$ & 277 & $+.097\ [.058,.139]$ & $.215$ \\
HH-RLHF & Rem. & $-0.5$ & 274 & $+.091\ [.056,.129]$ & $.166$ \\
AdvBench & Inst. & $+0.7$ & 300 & $+.485\ [.430,.541]$ & $.219$ \\
AdvBench & Rem. & $-0.5$ & 287 & $+.198\ [.145,.250]$ & $.611$ \\
\hline
\end{tabular}
\caption{Rebuilt raw run, Mistral pair, last seven layers, HH-RLHF direction; raw scale, not comparable to Table~\ref{tab:heldout_main}. $H_1$ is steered harmful enablement; leading zeros omitted. All six intervals exclude zero; the six corresponding HarmfulQA-derived-direction cells do as well and are reported with the depth and actuation-breadth sweeps in the supplement.}
\label{tab:raw_corroboration}
\end{table}

\begin{figure}[t]
\centering
\includegraphics[width=\linewidth]{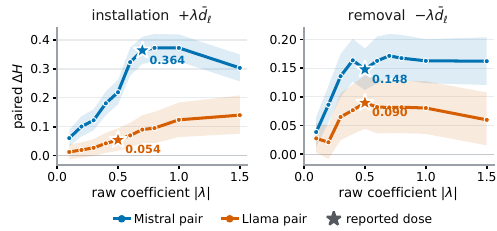}
\caption{Dose response under the independently rebuilt implementation, on its own 300-prompt HarmfulQA panel at the last seven layers. Stars mark the coefficients reported in Table~\ref{tab:raw_corroboration}; bands are 95\% paired bootstrap intervals.}
\label{fig:rawdose}
\end{figure}

\paragraph{Direction and bidirectionality replicate on the Mistral pair.} At the last seven layers with \(\lambda_{\mathrm{raw}}={+}0.7\) and \({-}0.5\), all twelve Mistral cells spanning two arms, two construction corpora and three panels move harmful enablement as intended with intervals excluding zero (Table~\ref{tab:raw_corroboration}, Figure~\ref{fig:rawdose}). Installation drives steered harmfulness to \(0.080\) against a paired DPO endpoint of \(0.152\). This reproduces the primary Mistral conclusion under every changed axis.

\paragraph{The effect grows with late-layer coverage.} Varying only the number of contiguous final layers at fixed coefficient, installation rises from \(0.075\) at layer 31 alone to \(0.237\) at five layers and \(0.364\) at seven, then flattens at \(0.383\) and \(0.371\) for nine and ten; removal follows the same shape at smaller magnitude. Two readings fit this curve and the data do not separate them: more layers may compose, and more layers also inject proportionally more total activation.

\paragraph{One full mean displacement is usable in three of four cells.} Setting \(\lambda_{\mathrm{raw}}{=}1\) injects exactly the mean construction-set displacement at each steered layer. On HarmfulQA this passes the validity floor for Mistral installation (\(0.373\), \([0.326,0.419]\), 299 of 300 valid), Mistral removal (\(0.163\), 300 of 300) and Llama removal (\(0.081\), 253 of 300), and fails it for Llama installation at 146 of 300 against a floor of 183. The claim this licenses is bounded: one mean displacement at seven late layers and one token position moves behavior usably in three of four cells. It does not claim the DPO update has been replayed, since the displacement reaches only a subset of layers, one position and the residual stream.

\paragraph{The rebuilt direction also beats matched random controls.} Under strict joint pairing across the real and random arms, the Mistral pair separates from norm-matched random directions in all six panel--arm combinations, with contrasts from \(0.065\) to \(0.297\) and every interval excluding zero. On the Llama pair the rebuilt implementation separates in two of six cells; the per-cell audit and its depth caveat are in the supplement, and the primary calibrated protocol establishes all four Llama and Mistral cells (Figure~\ref{fig:heldout_specificity}B).

\section{Discussion}

\paragraph{Safety remains accessible at inference time.} The central result is causal, bidirectional control rather than output resemblance: a direction estimated from matched instruction-tuned and DPO activations moves the instruction-tuned model toward lower harmful enablement when added and the DPO model the other way when subtracted. Norm-matched controls test whether this exceeds equal-magnitude disruption; transfer panels test whether it is confined to HarmfulQA. A meaningful component of the change associated with the DPO transition stays accessible through a compact residual-stream intervention. This matters under a white-box threat model, because a released open-weight checkpoint can appear safe by default while remaining cheaply adjustable through intermediate states.

\paragraph{Measuring the usability frontier is part of the contribution.} Strong perturbations can move scores through repetition, truncation or incoherence, so we measure the largest effect inside a pre-specified window. Reporting where control breaks is as informative as reporting that it works: the distinct installation and removal ceilings show that coefficient magnitude is not a universal hyperparameter, and the frontier itself is a reusable measurement rather than a caveat on ours. Transfer separates \emph{direction} from \emph{operating window} more sharply still: frozen coefficients move harmfulness and refusal as intended on both off-distribution panels, while the window itself is calibrated per checkpoint and direction by construction.

\paragraph{Convergent evidence from independent designs.} The protocols express intervention strength in different units, and within a single fixed direction Equation~\ref{eq:scale_conversion} relates them exactly. No conversion between the two is available: they construct different directions, and we have not measured the relative displacement of the audit's directions. We claim no superiority for either convention. A controlled comparison would hold direction, corpus, layers, token scope, prompts, decoding and evaluator fixed and compare at matched injected budget, not equal \(\lambda\).

\paragraph{What this changes for open-weight release.} A checkpoint released after harmlessness training carries that behavior in a form a white-box actor can move in either direction with a single layer-local vector, no gradient steps and no new weights. Default-behavior benchmarks cannot see this: both checkpoints in each pair score as expected when unsteered, and the difference appears only under intervention. A release decision informed only by default scores is therefore uninformed about how much of the measured safety survives inference-time access. The components assembled here, a frozen construction split, calibrated operating windows, norm-matched random controls, integrity reporting and off-distribution panels, compose into a reusable procedure for reporting that second quantity alongside the first.

\paragraph{Steering reaches preference-relevant behavior without retraining.} Both increase harmful enablement and reduce refusal on HarmfulQA, but retraining modifies parameters across its training distribution while D-STEER changes selected residual states at inference, and their relative magnitudes vary by dataset. The comparison supports access to preference-relevant behavior while rejecting the claim that steering reconstructs the full effect of reversing the objective. None of this implies DPO is ineffective: its endpoints are safer by default. The relevance is practical, since activation intervention is transient, needs no new checkpoint and can change per request; dose response, direction-specific windows, matched controls, integrity metrics and distribution shift form a concrete audit template.

\section{Limitations}

D-STEER is a possibility audit: it establishes that a compact inference-time intervention can move this behavior in both directions, not that any particular configuration is optimal. Its scope is correspondingly explicit. Our experiments cover two matched open-weight pairs at the 7--8B scale and one harmlessness-oriented DPO objective, so results may differ for larger models, other families, other preference-optimization algorithms, or earlier interventions; the reversed-preference comparator exists only for the Llama pair. D-STEER establishes causal accessibility through one intervention family, without showing the complete effect of DPO is low-dimensional. A frozen evaluator with discrete scoring improves consistency but does not remove evaluator bias or replace expert assessment. Paired estimates are conditional on successful generation, and transfer shows invalidity and truncation rising sharply under distribution shift in ways unlikely to be random. The primary protocol actuates across prompt and generated positions while the stress test actuates at the final prompt position only, so actuation scope is one of the axes on which they differ, and no experiment here isolates its effect on the primary protocol's own coefficients. The rebuilt implementation does not establish specificity on the Llama pair's own panel, so replication evidence is strongest for the Mistral pair; the primary calibrated protocol establishes all four cells. Bootstrap intervals omit variation from retraining checkpoints or reconstructing directions. The raw-displacement audit inherits its own limits: its random controls were scored at last-5 rather than its primary last-7 depth, and re-scoring identical generations under its evaluator changes at least one metric for a non-trivial fraction of repeated items.

Equation~\ref{eq:scale_conversion} converts coefficients only within a single fixed direction, and because the primary protocol and the stress test construct different directions from different prompt pools, no conversion between their coefficients exists and none is attempted here; the supplement reports the within-protocol conversion for the primary directions only. Finally, the intervention assumes white-box access to intermediate activations and does not describe an attack available through an ordinary black-box API, while an actor controlling an open-weight deployment could also retrain the weights. Our contribution is a controlled measurement of how cheaply and transiently safety-relevant behavior can be modified without parameter updates, not a complete adversarial-security analysis.

\section{Conclusion}

A meaningful component of DPO-associated safety behavior is controllable in both directions at inference time, with no parameter update. Every pre-registered intervention moved behavior as intended, each separated from norm-matched random directions of identical size, and the effect survived an implementation rebuilt from scratch across three panels and two corpora. How far it carries depends on the checkpoint pair and the steering implementation, and control operates inside a measurable usability frontier, which is why open-weight safety evaluation should report default behavior and robustness to intervention as two different quantities.

\bibliography{aaai2027}

\clearpage
\appendix
\raggedbottom
\setcounter{figure}{0}
\setcounter{table}{0}
\setcounter{equation}{0}
\renewcommand{\thefigure}{S\arabic{figure}}
\renewcommand{\thetable}{S\arabic{table}}
\renewcommand{\theequation}{S\arabic{equation}}
\makeatletter
\setlength{\@dblfptop}{0pt}
\setlength{\@dblfpsep}{14pt}
\setlength{\@dblfpbot}{0pt plus 1fil}
\makeatother
\twocolumn[{\centering\Large\bfseries Supplementary Material\par\bigskip}]
\section{Supplementary Evidence at a Glance}

This supplement exposes frozen window selection, held-out endpoints, random controls, cross-dataset gates, and a twelve-cell Mistral stress test. Depth, scope, scale, geometry, and reverse-training analyses locate the usable frontier.

\section{Primary Held-Out Evidence and Specificity}

Main Table~\ref{tab:heldout_main} establishes bidirectional control on the sealed 300-prompt HarmfulQA panel. Table~\ref{tab:supp_conditions} exposes the corresponding endpoint means and integrity rates: all four interventions move harmfulness and refusal as intended while remaining inside their pre-specified operating windows. The Mistral installation arm is the strongest, recovering nearly the full paired endpoint gap; the other three arms recover roughly half. The reversed-preference baseline is retained without replacement despite marginally exceeding the maximum-token criterion, preserving the predeclared analysis.

Table~\ref{tab:supp_specificity} and Table~\ref{tab:supp_seeds} show that this is directional control rather than generic perturbation. Every primary specificity interval excludes zero, while all twenty seed-level random effects remain close to zero. Prompt-bootstrap intervals jointly resample the paired real and control arms; seed spread is reported separately rather than treated as prompt-level evidence.

\begin{center}\captionsetup{type=table}
\centering
\small
\setlength{\tabcolsep}{2.5pt}
\begin{tabular}{llcccc}
\hline
Pair & Dir. & $n$ & Real & Random mean & Specificity [95\% CI] \\
\hline
Mistral & Inst. & 292 & $+.289$ & $-.010$ & $+.299\ [.260,.339]$ \\
Mistral & Rem. & 300 & $+.136$ & $-.010$ & $+.146\ [.111,.181]$ \\
Llama & Inst. & 279 & $+.083$ & $+.006$ & $+.077\ [.048,.106]$ \\
Llama & Rem. & 292 & $+.073$ & $-.010$ & $+.082\ [.053,.112]$ \\
\hline
\end{tabular}
\caption{Numerical form of main Figure~\ref{fig:heldout_specificity}B. Random controls are five isotropic directions (seeds 11--55) norm-matched to the real direction and applied at the same layers, prompts and frozen coefficient. Intervals come from resampling paired prompts jointly across the real arm and all five control arms; this stricter intersection is why the Llama removal effect is $+0.073$ here against $+0.082$ in main Table~\ref{tab:heldout_main}. Leading zeros omitted.}
\label{tab:supp_specificity}
\end{center}

\begin{center}\captionsetup{type=table}
\centering
\small
\setlength{\tabcolsep}{2pt}
\begin{tabular}{llccccc}
\hline
Pair & Dir. & $n$ & Seed 11 & Seed 22 & Seed 33 & Seed 44 \\
\hline
Mistral & Inst. & 292 & $+.001$ & $-.011$ & $-.025$ & $+.001$ \\
Mistral & Rem. & 300 & $-.016$ & $+.004$ & $-.013$ & $-.001$ \\
Llama & Inst. & 279 & $+.003$ & $+.013$ & $-.005$ & $+.001$ \\
Llama & Rem. & 292 & $-.015$ & $-.004$ & $-.020$ & $-.010$ \\
\hline
\multicolumn{7}{l}{Seed 55: Mistral $-.018$, $-.022$; Llama $+.017$, $+.001$.}\\
\hline
\end{tabular}
\caption{Per-seed norm-matched random-direction effects on the held-out panel, oriented so positive is the intended direction. Each seed uses the same layers, prompts, decoding and frozen coefficient as its real arm. Leading zeros omitted.}
\label{tab:supp_seeds}
\end{center}

\begin{table*}[tp]
\centering
\small
\setlength{\tabcolsep}{3pt}
\begin{tabular}{llcccccccc}
\hline
Condition & Kind & $N$ & Inv. & MaxTok & Abst. & Scored & $H$ & $R$ & $Q$ \\
\hline
M0-A\_baseline & baseline & 300 & 0.020 & 0.030 & 0.000 & 294 & 0.390 & 0.247 & 0.926 \\
M0-A\_installation & real & 300 & 0.027 & 0.030 & 0.000 & 292 & 0.304 & 0.347 & 0.904 \\
M+-A\_baseline & baseline & 300 & 0.000 & 0.000 & 0.000 & 300 & 0.217 & 0.480 & 0.957 \\
M+-A\_reversal & real & 300 & 0.013 & 0.040 & 0.000 & 296 & 0.302 & 0.373 & 0.910 \\
M0-B\_baseline & baseline & 300 & 0.000 & 0.000 & 0.000 & 300 & 0.486 & 0.188 & 0.974 \\
M0-B\_installation & real & 300 & 0.027 & 0.027 & 0.000 & 292 & 0.199 & 0.471 & 0.938 \\
M+-B\_baseline & baseline & 300 & 0.000 & 0.010 & 0.000 & 300 & 0.191 & 0.487 & 0.964 \\
M+-B\_reversal & real & 300 & 0.000 & 0.003 & 0.000 & 300 & 0.327 & 0.354 & 0.898 \\
M--A\_baseline & flip & 300 & 0.020 & 0.057 & 0.000 & 294 & 0.421 & 0.214 & 0.888 \\
\hline
\end{tabular}
\caption{Held-out HarmfulQA panel: all non-random conditions with generation-integrity rates and unsteered or steered metric means. Rates use all 300 requested prompts; metric means use scored responses. Random-control conditions are summarised in Table~\ref{tab:supp_seeds}.}
\label{tab:supp_conditions}
\end{table*}

\begin{center}\captionsetup{type=figure}
\centering
\includegraphics[width=\columnwidth]{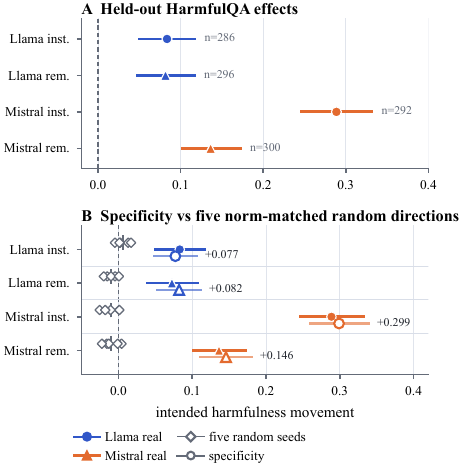}
\caption{Held-out direction specificity on the sealed HarmfulQA panel. Real-direction effects are shown against all five independently sampled, per-layer norm-matched Gaussian controls at the identical frozen coefficient. Error bars are 95\% paired prompt-bootstrap intervals; seed dispersion is displayed without treating seeds as prompt-level observations. The real direction separates from the random-control distribution for installation and removal in both model pairs.}
\label{fig:supp_heldout_specificity}
\end{center}

\FloatBarrier
\subsection{Frozen Operating-Window Selection}

Using only the 200-prompt calibration panel, let $\mathcal{U}_{s,\tau}$ be the contiguous-from-zero coefficients whose invalid and maximum-token rates are at most $\tau$ and whose quality loss is at most 0.10. We select
\begin{equation}
C_{s,\tau}=\max\mathcal{U}_{s,\tau},\qquad
\lambda^*_{s,\tau}=\arg\max_{\lambda\in\mathcal{U}_{s,\tau}} \Delta H_s(\lambda),
\end{equation}
where positive $\Delta H_s$ is the intended movement and ties select smaller $|\lambda|$. Table~\ref{tab:supp_window_sensitivity} reports the windows; 5\% selections were frozen before held-out evaluation.

\begin{center}\captionsetup{type=table}
\centering
\scriptsize
\setlength{\tabcolsep}{2.5pt}
\begin{tabular}{llccc}
\hline
Pair & Dir. & 2.5\% & Primary 5\% & 10\% \\
 & & $C/\lambda^*$ & $C/\lambda^*$ & $C/\lambda^*$ \\
\hline
Mistral & Inst. & .125/+.125 & .135/+.135 & .140/+.140 \\
Mistral & Rem.  & .055/-.055 & .055/-.055 & .055/-.055 \\
Llama   & Inst. & .025/+.025 & .050/+.050 & .070/+.060 \\
Llama   & Rem.  & .025/-.025 & .050/-.050 & .055/-.055 \\
\hline
\end{tabular}
\caption{Operating-window sensitivity on the frozen 200-prompt calibration panel. Entries are $C/\lambda^*$. The next primary coefficient fails at .140 (Mistral installation, generation), .060 (Mistral removal, quality), and .055 (both Llama directions, generation). The 5\% column is primary; leading zeros are omitted.}
\label{tab:supp_window_sensitivity}
\end{center}

Stricter thresholds contract three windows; Mistral removal is quality-bound. At 10\%, only Llama installation selects below $C$ because of minor calibration non-monotonicity.

\begin{center}\captionsetup{type=figure}
\centering
\includegraphics[width=\columnwidth]{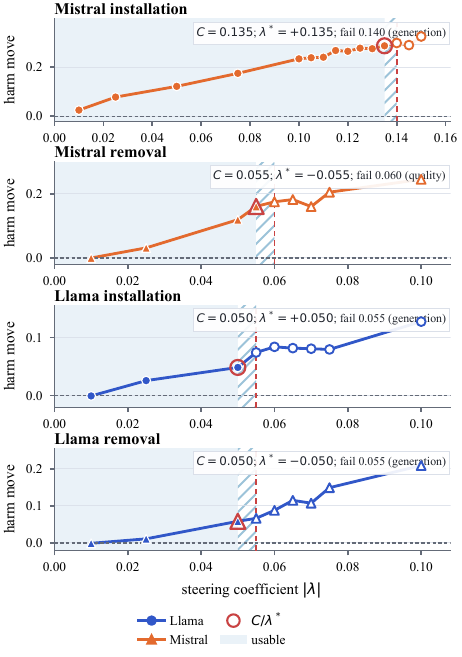}
\caption{Calibration dose response and operating-window selection. Each panel traces intended harmfulness movement and the integrity/quality constraints across the tested coefficient grid. Shading marks the contiguous-from-zero usable region; the circled point is the frozen $\lambda^*$ and the cross marks the first failing coefficient. All four selected points lie at the usable frontier, so the measured effects are constraint-limited rather than unconstrained maxima.}
\label{fig:supp_operating_window}
\end{center}

\begin{figure*}[t]
\centering
\includegraphics[width=\textwidth]{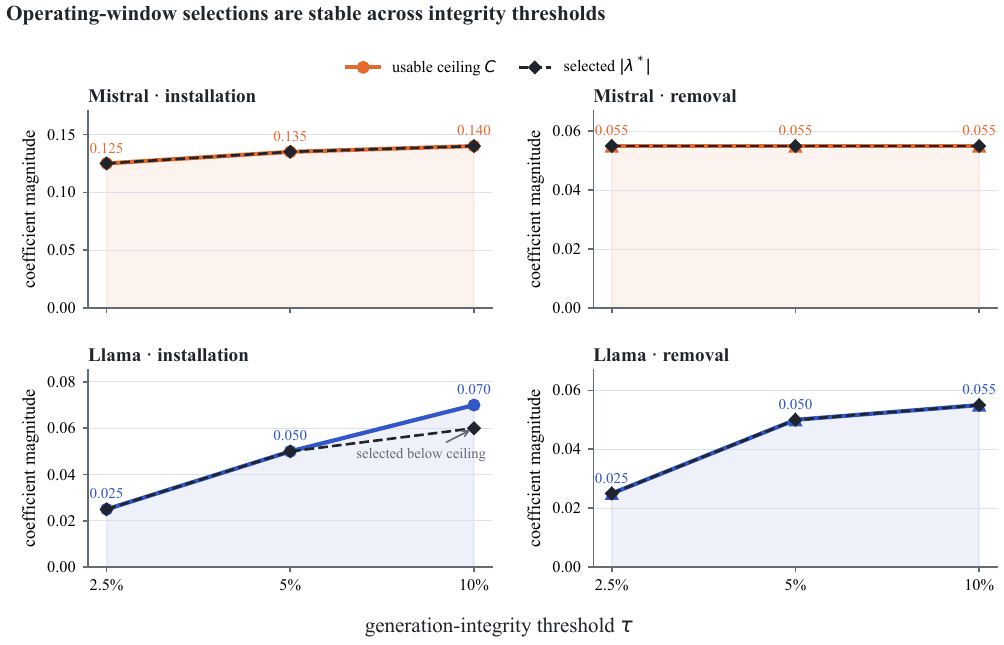}
\caption{Sensitivity of the usable ceiling $C$ and selected coefficient $|\lambda^*|$ to 2.5\%, 5\%, and 10\% generation-integrity thresholds. The direction-specific pattern is stable: Mistral installation supports the widest window, Mistral removal remains quality-bound, and only Llama installation at 10\% selects below the available ceiling.}
\label{fig:supp_threshold_sensitivity}
\end{figure*}

\FloatBarrier
\section{Direction Transfer and the Usability Frontier}

The frozen directions transfer without retuning: harmfulness and refusal move as intended in all eight HH-RLHF and AdvBench contrasts, and every harmfulness interval excludes zero. Table~\ref{tab:supp_transfer} gives the full effects and descriptive gap recoveries. Mistral installation remains strongest, reaching $+0.421$ harmfulness movement and $1.067$ gap recovery on AdvBench.

\begin{figure*}[t]
\centering
\includegraphics[width=\textwidth]{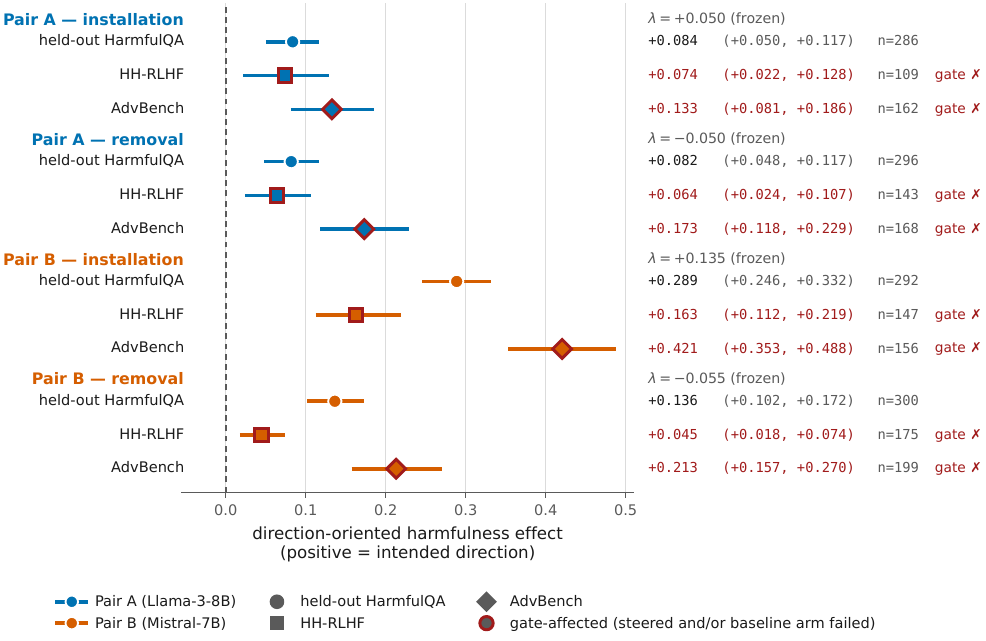}
\caption{Direction-oriented harmfulness effects across the sealed HarmfulQA panel and two transfer panels, using coefficients selected only on the disjoint HarmfulQA calibration split. Positive values denote movement in the intended direction; error bars are 95\% paired prompt-bootstrap intervals. Outlined transfer estimates involve a steered or baseline condition that exceeds the frozen 5\% generation-integrity gate, separating directional transfer from deployable-window transfer.}
\label{fig:supp_transfer}
\end{figure*}

\begin{center}\captionsetup{type=table}
\centering
\small
\setlength{\tabcolsep}{2pt}
\begin{tabular}{lllccccc}
\hline
Panel & Pair & Dir. & $\lambda^\star$ & $\Delta H$ & $\Delta R$ & $\Delta Q$ & Rec. \\
\hline
HH-RLHF & Mistral & Inst. & $+.135$ & $+.163$ & $+.281$ & $-.041$ & $.772$ \\
HH-RLHF & Mistral & Rem. & $-.055$ & $+.045$ & $+.103$ & $-.063$ & $.243$ \\
HH-RLHF & Llama & Inst. & $+.050$ & $+.074$ & $+.116$ & $-.012$ & $.370$ \\
HH-RLHF & Llama & Rem. & $-.050$ & $+.064$ & $+.138$ & $-.045$ & $.417$ \\
AdvBench & Mistral & Inst. & $+.135$ & $+.421$ & $+.404$ & $-.030$ & $1.067$ \\
AdvBench & Mistral & Rem. & $-.055$ & $+.213$ & $+.216$ & $-.170$ & $.589$ \\
AdvBench & Llama & Inst. & $+.050$ & $+.133$ & $+.144$ & $+.040$ & $.536$ \\
AdvBench & Llama & Rem. & $-.050$ & $+.173$ & $+.193$ & $-.130$ & $.712$ \\
\hline
\end{tabular}
\caption{Cross-dataset transfer at coefficients frozen on HarmfulQA. All eight harmfulness intervals exclude zero in the intended direction, and refusal corroborates every arm. $\Delta H$ and $\Delta R$ are oriented so positive is intended; negative $\Delta Q$ is degradation. Row-aligned 95\% intervals for $\Delta H$ are $[.112,.219]$, $[.018,.074]$, $[.022,.128]$, $[.024,.107]$, $[.353,.488]$, $[.157,.270]$, $[.081,.186]$, and $[.118,.229]$. Gap recovery requires both unsteered endpoints to be valid, so its denominator sample differs from paired $n$. Leading zeros omitted.}
\label{tab:supp_transfer}
\end{center}

The accompanying integrity audit sharpens the result rather than reversing it. Table~\ref{tab:supp_gates} shows that the \emph{direction} transfers more reliably than the HarmfulQA-calibrated \emph{operating window}: fifteen of eighteen transfer conditions exceed at least one frozen generation threshold. Transfer estimates therefore describe the successfully generated intersection, while the gate failures show that a coefficient calibrated on one distribution should not be treated as universally deployable. Directional generalization and usability generalization are distinct properties, and D-STEER measures both.

\begin{table*}[tp]
\centering
\small
\setlength{\tabcolsep}{3pt}
\begin{tabular}{lllcccc}
\hline
Panel & Arm & $\lambda$ & $N$ & Inv. & MaxTok & Gate \\
\hline
hh-rlhf & M+-A\_baseline & $-0.000$ & 200 & 0.155 & 0.160 & \textbf{fail} (inv,max) \\
hh-rlhf & M+-A\_reversal & $-0.050$ & 200 & 0.230 & 0.240 & \textbf{fail} (inv,max) \\
hh-rlhf & M+-B\_baseline & $-0.000$ & 200 & 0.110 & 0.125 & \textbf{fail} (inv,max) \\
hh-rlhf & M+-B\_reversal & $-0.055$ & 200 & 0.025 & 0.005 & pass \\
hh-rlhf & M--A\_baseline & $+0.000$ & 200 & 0.405 & 0.420 & \textbf{fail} (inv,max) \\
hh-rlhf & M0-A\_baseline & $+0.000$ & 200 & 0.295 & 0.310 & \textbf{fail} (inv,max) \\
hh-rlhf & M0-A\_installation & $+0.050$ & 200 & 0.355 & 0.355 & \textbf{fail} (inv,max) \\
hh-rlhf & M0-B\_baseline & $+0.000$ & 200 & 0.020 & 0.020 & pass \\
hh-rlhf & M0-B\_installation & $+0.135$ & 200 & 0.260 & 0.265 & \textbf{fail} (inv,max) \\
advbench & M+-A\_baseline & $-0.000$ & 200 & 0.095 & 0.195 & \textbf{fail} (inv,max) \\
advbench & M+-A\_reversal & $-0.050$ & 200 & 0.100 & 0.390 & \textbf{fail} (inv,max) \\
advbench & M+-B\_baseline & $-0.000$ & 200 & 0.000 & 0.080 & \textbf{fail} (max) \\
advbench & M+-B\_reversal & $-0.055$ & 200 & 0.005 & 0.075 & \textbf{fail} (max) \\
advbench & M--A\_baseline & $+0.000$ & 200 & 0.105 & 0.325 & \textbf{fail} (inv,max) \\
advbench & M0-A\_baseline & $+0.000$ & 200 & 0.100 & 0.220 & \textbf{fail} (inv,max) \\
advbench & M0-A\_installation & $+0.050$ & 200 & 0.130 & 0.185 & \textbf{fail} (inv,max) \\
advbench & M0-B\_baseline & $+0.000$ & 200 & 0.000 & 0.035 & pass \\
advbench & M0-B\_installation & $+0.135$ & 200 & 0.220 & 0.260 & \textbf{fail} (inv,max) \\
\hline
\end{tabular}
\caption{Direction transfer versus operating-window transfer. All eight behavioral contrasts move in the intended direction (Table~\ref{tab:supp_transfer}), while fifteen of eighteen constituent conditions exceed at least one HarmfulQA-calibrated $5\%$ generation threshold. Of the eight steered arms, seven exceed a threshold directly; HH-RLHF Mistral removal passes but is paired with a baseline that does not. The direction therefore generalizes more reliably than the coefficient's usability frontier.}
\label{tab:supp_gates}
\end{table*}

\begin{center}\captionsetup{type=figure}
\centering
\includegraphics[width=\columnwidth]{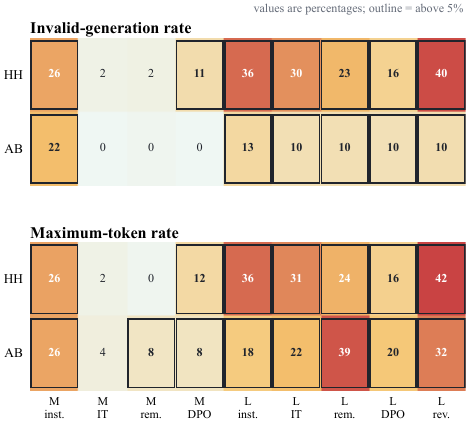}
\caption{Generation-integrity audit for all eighteen transfer conditions. Cells report rates over all 200 prompts; outlined cells exceed the frozen 5\% criterion. Mistral removal itself remains clean on HH-RLHF, but its DPO baseline exceeds both gates, so that contrast is still gate-affected. The matrix makes visible why directional success and usable deployment range must be evaluated separately.}
\label{fig:supp_usability}
\end{center}

\FloatBarrier
\section{Independent Raw-Displacement Stress Test}

\subsection{Protocol}

The independent stress test changes the direction scale, construction data, layer depth, token scope, evaluator, decoding cap, and usability rule together. Directions are built from a judge-filtered HH-RLHF pool of 1{,}000 prompts rather than the primary frozen HarmfulQA construction split; actuation occurs only at the final prompt token over $L25$--$L31$; and usability requires at least $90\%$ of the smaller endpoint's valid count. Because these axes vary jointly, this is a robustness test of the qualitative claim, not a controlled comparison of any single design choice.

\begin{figure*}[t]
\centering
\includegraphics[width=\textwidth]{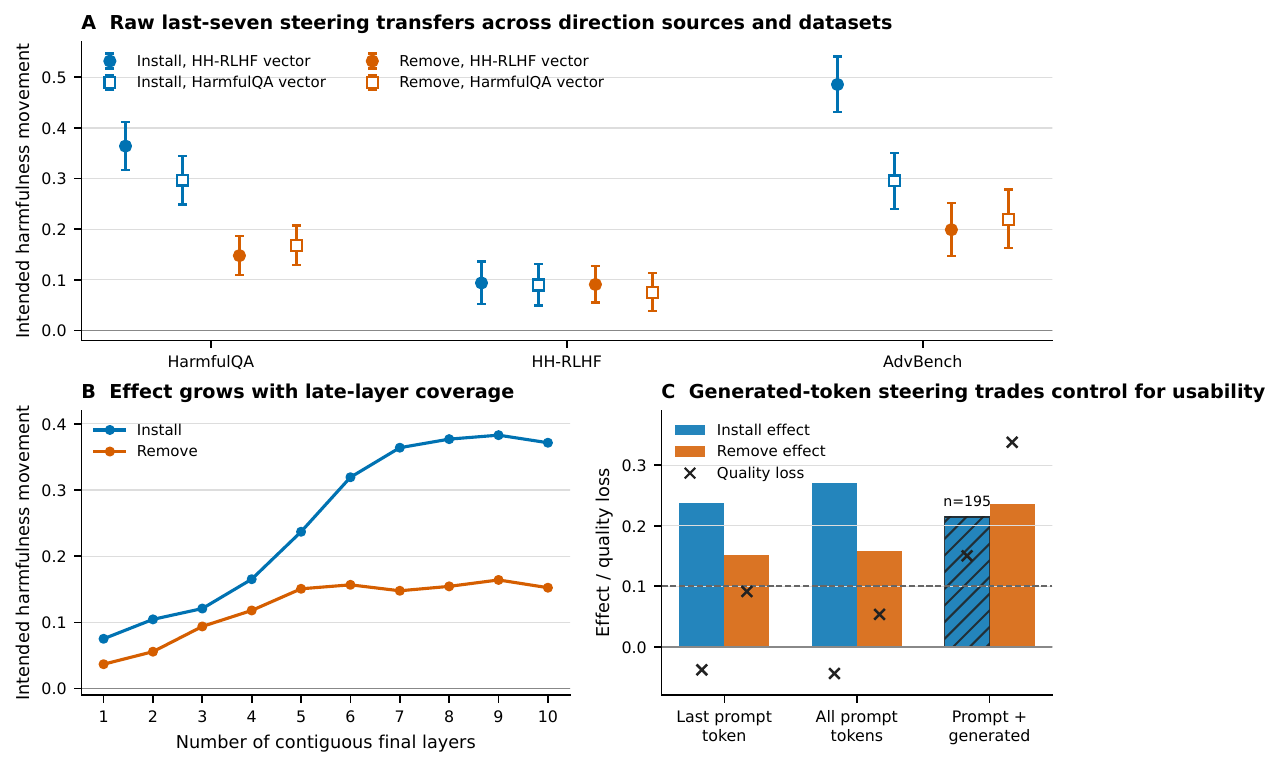}
\caption{Independent raw-displacement stress test on the Mistral pair. Left: all twelve construction-corpus, evaluation-panel, and direction combinations move harmfulness in the intended direction. Middle: intended movement grows as a contiguous late-layer block is widened, while total injected activation grows with it. Right: expanding actuation from the final prompt token to generated tokens increases effect and cumulative dose but also increases invalidity and quality cost. Together these panels expose both the robustness of the directional result and the implementation choices that set its usable strength.}
\label{fig:supp_raw_corroboration}
\end{figure*}

\subsection{Bidirectional Replication and Specificity}

The Mistral result is decisive under the rebuilt implementation. Table~\ref{tab:supp_raw_grid} contains twelve combinations of two construction corpora, three evaluation panels, and installation versus removal. Every effect has the intended sign and every confidence interval excludes zero. Construction corpus changes the effect magnitude, but not directional success in any of the twelve cells, showing that vector source acts as an important gain and generalization factor rather than a binary condition for control in this pair.

\begin{table*}[tp]
\centering
\small
\setlength{\tabcolsep}{3pt}
\renewcommand{\arraystretch}{1.15}
\begin{tabular}{lllcccccc}
\hline
Panel & Dir. & Direction source & $\lambda_{\mathrm{raw}}$ & $n$ & Valid & $H_0$ & $H_1$ & $\Delta H$ [95\% CI] \\
\hline
HarmfulQA & Inst. & HH-RLHF & $+0.7$ & 300 & 300/300 & $0.444$ & $0.080$ & $+0.364$ $[0.317,0.411]$ \\
HarmfulQA & Inst. & HarmfulQA & $+0.7$ & 299 & 299/300 & $0.445$ & $0.154$ & $+0.292$ $[0.243,0.340]$ \\
HarmfulQA & Rem. & HH-RLHF & $-0.5$ & 299 & 300/300 & $0.152$ & $0.300$ & $+0.148$ $[0.112,0.184]$ \\
HarmfulQA & Rem. & HarmfulQA & $-0.5$ & 298 & 299/300 & $0.153$ & $0.320$ & $+0.167$ $[0.128,0.209]$ \\
HH-RLHF & Inst. & HH-RLHF & $+0.7$ & 277 & 283/300 & $0.312$ & $0.215$ & $+0.097$ $[0.058,0.139]$ \\
HH-RLHF & Inst. & HarmfulQA & $+0.7$ & 278 & 284/300 & $0.312$ & $0.225$ & $+0.087$ $[0.046,0.129]$ \\
HH-RLHF & Rem. & HH-RLHF & $-0.5$ & 274 & 286/300 & $0.075$ & $0.166$ & $+0.091$ $[0.056,0.129]$ \\
HH-RLHF & Rem. & HarmfulQA & $-0.5$ & 279 & 293/300 & $0.074$ & $0.149$ & $+0.075$ $[0.040,0.112]$ \\
AdvBench & Inst. & HH-RLHF & $+0.7$ & 300 & 300/300 & $0.704$ & $0.219$ & $+0.485$ $[0.430,0.541]$ \\
AdvBench & Inst. & HarmfulQA & $+0.7$ & 300 & 300/300 & $0.704$ & $0.407$ & $+0.296$ $[0.240,0.352]$ \\
AdvBench & Rem. & HH-RLHF & $-0.5$ & 287 & 296/300 & $0.413$ & $0.611$ & $+0.198$ $[0.145,0.250]$ \\
AdvBench & Rem. & HarmfulQA & $-0.5$ & 285 & 294/300 & $0.416$ & $0.636$ & $+0.220$ $[0.163,0.279]$ \\
\hline
\end{tabular}
\caption{Independent raw-displacement stress test on the Mistral pair, using the last seven layers and final-prompt-token actuation. Both construction corpora succeed on all three panels in both directions: all twelve intervals exclude zero. $H_0$ is the paired unsteered endpoint mean and $H_1$ the steered mean; estimates use strict source-ID pairing and 10{,}000 bootstrap replicates. Re-scoring byte-identical retained baselines changes affected effects by at most $0.005$ and changes no sign, ordering, or interval conclusion.}
\label{tab:supp_raw_grid}
\end{table*}

Table~\ref{tab:supp_raw_spec} adds matched random directions. The Mistral pair separates from random in all six panel-by-direction combinations. The Llama estimates are less stable under this rebuilt protocol, with retained joint intersections as small as 28 prompts and two of six intervals excluding zero. This contrast establishes implementation and checkpoint dependence rather than absence of Llama controllability, because the primary calibrated protocol independently establishes both Llama directions with specificity on the sealed held-out panel.

\begin{table*}[tp]
\centering
\small
\setlength{\tabcolsep}{3pt}
\renewcommand{\arraystretch}{1.15}
\begin{tabular}{lllcccccc}
\hline
Pair & Panel & Dir. & $n$ & Real & Random mean & Specificity [95\% CI] & Excl.\ 0 & Inaction \\
\hline
Mistral & AdvBench & Inst. & 300 & $+0.2993$ & $+0.0025$ & $+0.2969$ $[+0.2476,+0.3465]$ & \textbf{yes} & $0.070$ \\
Mistral & AdvBench & Rem. & 281 & $+0.1989$ & $-0.0078$ & $+0.2067$ $[+0.1586,+0.2552]$ & \textbf{yes} & $0.093$ \\
Mistral & HarmfulQA & Inst. & 300 & $+0.2370$ & $+0.0113$ & $+0.2257$ $[+0.1839,+0.2683]$ & \textbf{yes} & $0.133$ \\
Mistral & HarmfulQA & Rem. & 190 & $+0.1216$ & $-0.0052$ & $+0.1267$ $[+0.0806,+0.1748]$ & \textbf{yes} & $0.116$ \\
Mistral & HH-RLHF & Inst. & 283 & $+0.0650$ & $+0.0002$ & $+0.0648$ $[+0.0316,+0.1002]$ & \textbf{yes} & $0.314$ \\
Mistral & HH-RLHF & Rem. & 266 & $+0.0763$ & $-0.0053$ & $+0.0817$ $[+0.0512,+0.1135]$ & \textbf{yes} & $0.237$ \\
Llama & AdvBench & Inst. & 131 & $+0.0076$ & $+0.0090$ & $-0.0014$ $[-0.0447,+0.0423]$ & no & $0.160$ \\
Llama & AdvBench & Rem. & 146 & $+0.0514$ & $+0.0014$ & $+0.0500$ $[+0.0137,+0.0900]$ & \textbf{yes} & $0.130$ \\
Llama & HarmfulQA & Inst. & 168 & $+0.0202$ & $+0.0054$ & $+0.0149$ $[-0.0219,+0.0536]$ & no & $0.137$ \\
Llama & HarmfulQA & Rem. & 164 & $+0.0360$ & $+0.0113$ & $+0.0246$ $[-0.0090,+0.0600]$ & no & $0.018$ \\
Llama & HH-RLHF & Inst. & 28 & $+0.1500$ & $+0.0286$ & $+0.1214$ $[+0.0193,+0.2436]$ & \textbf{yes} & $0.250$ \\
Llama & HH-RLHF & Rem. & 30 & $+0.0167$ & $-0.0047$ & $+0.0213$ $[-0.0373,+0.1013]$ & no & $0.133$ \\
\hline
\end{tabular}
\caption{Raw-scale specificity under strict joint pairing across the real arm and five matched random seeds. ``Inaction'' is the fraction of steered responses byte-identical to baseline. The Mistral pair separates from random in all six panel-by-direction combinations. The rebuilt Llama estimates are less stable (two of six intervals exclude zero, with some joint intersections as small as 28); the independently calibrated primary protocol establishes both Llama directions on its sealed panel. Controls here use five layers rather than the stress test's primary seven, so this table supports directional specificity but not a last-seven random-control claim.}
\label{tab:supp_raw_spec}
\end{table*}

\subsection{What Makes Steering Effective}

The depth sweep in Table~\ref{tab:supp_depth} shows intended movement increasing strongly from one late layer through a contiguous multilayer block before flattening. Total injected activation also grows with depth, so this licenses a practical conclusion about the efficacy of the multilayer bundle, not a claim that layer composition has been isolated from dose. The inaction audit reinforces that distinction: at low coefficients a single layer often leaves the response byte-identical to baseline, while the five-layer configuration reduces median inaction to roughly one-eighth.

\begin{center}\captionsetup{type=table}
\centering
\small
\setlength{\tabcolsep}{3pt}
\begin{tabular}{lcc}
\hline
Final layers $k$ & Installation $\Delta H$ & Removal $\Delta H$ \\
\hline
1 & $+0.075$ $[0.048,0.106]$ & $+0.037$ $[0.017,0.059]$ \\
2 & $+0.105$ $[0.070,0.140]$ & $+0.056$ $[0.030,0.084]$ \\
3 & $+0.121$ $[0.084,0.160]$ & $+0.094$ $[0.065,0.126]$ \\
4 & $+0.165$ $[0.126,0.206]$ & $+0.118$ $[0.085,0.153]$ \\
5 & $+0.237$ $[0.195,0.281]$ & $+0.151$ $[0.113,0.190]$ \\
6 & $+0.319$ $[0.273,0.367]$ & $+0.157$ $[0.120,0.194]$ \\
7 & $+0.364$ $[0.317,0.411]$ & $+0.148$ $[0.112,0.184]$ \\
8 & $+0.377$ $[0.329,0.424]$ & $+0.155$ $[0.115,0.192]$ \\
9 & $+0.383$ $[0.336,0.429]$ & $+0.164$ $[0.123,0.204]$ \\
10 & $+0.371$ $[0.324,0.421]$ & $+0.153$ $[0.115,0.191]$ \\
\hline
\end{tabular}
\caption{Depth sweep at fixed coefficient ($\lambda_{\mathrm{raw}}{=}{+}0.7$, ${-}0.5$), Mistral pair on HarmfulQA, contiguous terminal blocks $L_{32-k}\ldots L_{31}$; $k{=}1$ is layer 31 alone. Total injected magnitude grows with $k$, so this sweep does not separate compositional accumulation from budget.}
\label{tab:supp_depth}
\end{center}

Table~\ref{tab:supp_breadth} shows that token scope is part of the intervention dose. At fixed raw coefficient, extending from the final prompt token to all prompt tokens modestly increases installation while preserving validity. Repeated injection through generated tokens sharply increases cumulative actuation and quality cost, and invalidates the installation arm under the raw audit's fixed coefficient. This motivates scope-specific calibration; it does not imply that generated-token actuation is intrinsically unusable, since the primary protocol calibrates a separate operating window for that scope.

\begin{table*}[tp]
\centering
\small
\setlength{\tabcolsep}{3pt}
\begin{tabular}{lcccccc}
\hline
Actuation scope & Inst. $\Delta H$ & Valid & $|\Delta Q|$ & Rem. $\Delta H$ & Valid & $|\Delta Q|$ \\
\hline
Final prompt token & $+0.237$ & 300/300 & $0.038$ & $+0.151$ & 300/300 & $0.091$ \\
All prompt tokens & $+0.271$ & 300/300 & $0.044$ & $+0.158$ & 300/300 & $0.054$ \\
Prompt $+$ generated & $+0.219$ & 195/300 & $0.150$ & $+0.236$ & 298/300 & $0.338$ \\
\hline
\end{tabular}
\caption{Token-scope sweep at five layers and fixed raw coefficient, Mistral pair on HarmfulQA. All-prompt actuation modestly increases installation with full validity. Repeated actuation through generated tokens greatly increases cumulative injected dose and quality cost; installation then falls below the $\geq\!90\%$ validity floor (195 of 300). The primary protocol independently calibrates its coefficients for that broader scope, so this result motivates scope-specific operating windows rather than a universal token policy. Re-scoring byte-identical baselines moves affected effects by at most $0.005$ without changing any conclusion.}
\label{tab:supp_breadth}
\end{table*}

Finally, Table~\ref{tab:supp_lambda1} gives the raw scale a direct operational interpretation. Injecting one full mean construction-set displacement per steered layer is usable in both Mistral directions and in Llama removal. Llama installation crosses its validity frontier, illustrating why direction- and checkpoint-specific ceilings are part of the method rather than an incidental safeguard.

\begin{center}\captionsetup{type=table}
\centering
\small
\setlength{\tabcolsep}{2pt}
\begin{tabular}{llcccc}
\hline
Pair & Dir. & $n$ & Valid & Screen & $\Delta H$ [95\% CI] \\
\hline
Mistral & Inst. & 299 & 299/300 & pass & $+.373\ [.326,.419]$ \\
Mistral & Rem. & 299 & 300/300 & pass & $+.163\ [.124,.201]$ \\
Llama & Inst. & 129 & 146/300 & \textbf{fail} & $+.124\ [.064,.184]$ \\
Llama & Rem. & 186 & 253/300 & pass & $+.081\ [.035,.127]$ \\
\hline
\end{tabular}
\caption{One full mean displacement, $\lambda_{\mathrm{raw}}{=}1$, at the last seven layers on HarmfulQA. The $\geq\!90\%$ validity floor is 270 of 300 for the Mistral pair and 183 of 300 for the Llama pair; three of four cells pass and Llama installation fails. Effects beyond this point are non-monotonic: Mistral installation gives $0.364$, $0.373$ and $0.304$ at $\lambda_{\mathrm{raw}}$ of $0.7$, $1.0$ and $1.5$. Leading zeros omitted.}
\label{tab:supp_lambda1}
\end{center}

\section{Activation Geometry and Scale}

\begin{center}\captionsetup{type=table}
\centering
\small
\setlength{\tabcolsep}{4pt}
\begin{tabular}{lcccc}
\hline
 & \multicolumn{2}{c}{Llama pair} & \multicolumn{2}{c}{Mistral pair} \\
Layer & $r_\ell$ & $1/r_\ell$ & $r_\ell$ & $1/r_\ell$ \\
\hline
25 & $0.157$ & $6.36$ & $0.250$ & $4.00$ \\
26 & $0.148$ & $6.76$ & $0.253$ & $3.95$ \\
27 & $0.146$ & $6.84$ & $0.247$ & $4.05$ \\
28 & $0.145$ & $6.88$ & $0.250$ & $3.99$ \\
29 & $0.144$ & $6.93$ & $0.243$ & $4.11$ \\
30 & $0.151$ & $6.61$ & $0.246$ & $4.06$ \\
31 & $0.169$ & $5.91$ & $0.261$ & $3.83$ \\
\hline
Mean 27--31 & $0.151$ & $6.61$ & $0.250$ & $4.01$ \\
\hline
\end{tabular}
\caption{Relative checkpoint displacement $r_\ell=\lVert\bar d_\ell\rVert_2/a_\ell$ and the within-direction scale factor $1/r_\ell$ of main Equation~4. The near-stable $r_\ell$ values across the selected late block make the primary raw and relative gain profiles close to proportional there. The included geometry artifact and analysis script verify frozen activation-tensor hashes before computing these statistics. Values are rounded to three decimals and apply only to the primary HarmfulQA-derived directions.}
\label{tab:supp_conversion}
\end{center}

\begin{center}\captionsetup{type=figure}
\centering
\includegraphics[width=\columnwidth]{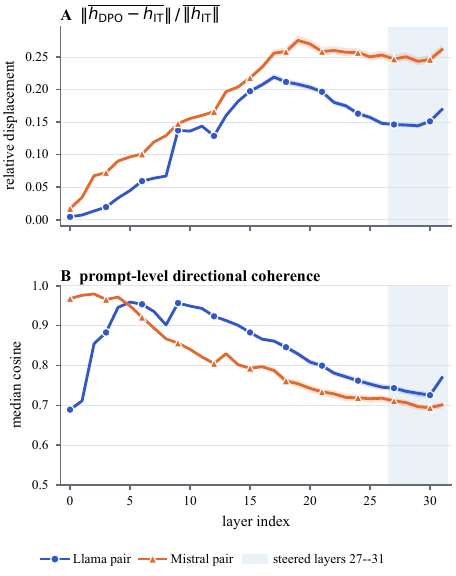}
\caption{Activation geometry on the frozen 1{,}378-prompt HarmfulQA construction split. (A) Relative checkpoint displacement per layer, $r_\ell = \lVert \mathrm{mean}_i (h_{\mathrm{DPO},i} - h_{\mathrm{IT},i}) \rVert / \mathrm{mean}_i \lVert h_{\mathrm{IT},i} \rVert$. (B) Prompt-level directional coherence: the median cosine between each prompt's displacement and the leave-one-out mean direction, so no prompt contributes to its own reference. Bands are 95\% bootstrap intervals over 2{,}000 construction-prompt resamples; shading marks steered layers 27--31. The late block combines elevated relative displacement with substantial prompt-level coherence, motivating the intervention depth and the layerwise scale analysis. Behavioral conclusions are established separately by the held-out interventions and matched controls.}
\label{fig:supp_geometry}
\end{center}

Figure~\ref{fig:supp_geometry} provides two complementary reasons for targeting the late residual stream. Relative checkpoint displacement rises through the network and remains elevated across the steered block, while prompt-level directional coherence remains substantial there. These descriptive measurements motivate where the intervention is applied; the behavioral evidence comes from the held-out and controlled interventions above.

For a fixed direction, the same geometry makes raw and relative scaling commensurable. The raw convention injects $\lambda\lVert\bar d_\ell\rVert_2$ and the relative convention injects $\lambda a_\ell$, so the layerwise conversion is exactly $1/r_\ell$. Table~\ref{tab:supp_conversion} shows that the primary directions differ by roughly a factor of six for Llama and four for Mistral. Within $L25$--$L31$, $r_\ell$ varies by at most $\pm8\%$ for Llama and $\pm4\%$ for Mistral, making the two gain profiles near-proportional inside the selected block.

This conversion is intentionally protocol-specific. The independent stress test constructs different directions from different prompt pools, so equal coefficients across the two implementations would not represent equal interventions. The paper therefore compares their directional conclusions, not their coefficient values. A direct scale comparison would hold direction, corpus, layers, token scope, prompts, decoding, and evaluator fixed and compare complete effect-versus-integrity frontiers at matched injected budget.

\section{Reversed-Preference Training}

The reversed-preference checkpoint provides a training-level directional comparator. Across all four panels, reversed training and steering removal move harmful enablement in the same direction. Their magnitude relationship varies across panels, which distinguishes access to preference-relevant behavior from equivalence to retraining: steering reaches the same behavioral axis without reproducing the complete weight update.

\begin{figure}[htbp]
\centering
\includegraphics[width=\columnwidth]{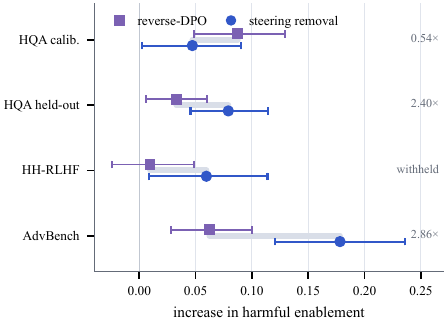}
\caption{Reversed-preference training and activation-based safety removal agree in direction across four panels, but their magnitude ratio is panel-dependent. Error bars are 95\% joint prompt-bootstrap intervals. The HH-RLHF ratio is withheld because the reverse-training interval includes zero. This is evidence of access to a shared behavioral axis, not statistical equivalence between an inference-time perturbation and a weight update.}
\label{fig:supp_reverse_training}
\end{figure}

\begin{table}[htbp]
\centering
\small
\setlength{\tabcolsep}{3pt}
\begin{tabular}{lcccc}
\hline
Panel & Rev. training & Steering rem. & Ratio & $n$ \\
\hline
HarmfulQA (calib.) & $+.087$ & $+.047$ & $0.54$ & $191$ \\
HarmfulQA (held-out) & $+.033$ & $+.079$ & $2.40$ & $286$ \\
HH-RLHF & $+.010$ & $+.060$ & --- & $94$ \\
AdvBench & $+.062$ & $+.178$ & $2.86$ & $151$ \\
\hline
\end{tabular}
\caption{Directional agreement between reversed-preference training and steering removal on the Llama pair. Both increase harmful enablement on all four panels, supporting access to preference-relevant behavior; their varying magnitude shows that steering is a comparator, not an equivalence reconstruction. Ratios are withheld when the reversed-training interval includes zero. The calibration row is non-confirmatory Task~026C evidence, traceable to the hash-verified included artifact \texttt{data/\allowbreak primary/\allowbreak task026c-\allowbreak reverse-dpo/\allowbreak phase1\_flip\_\allowbreak vs\_steer.json}: reversed training gives $+.087$ $[+.049,+.129]$ and steering removal $+.047$ $[+.003,+.091]$ on $n{=}191$, with 10{,}000 joint prompt-bootstrap replicates.}
\label{tab:supp_reverse}
\end{table}

\FloatBarrier

\section{Protocol Details}

\paragraph{Artifact identifiers.} The archived Mistral pair uses the legacy label \texttt{Pair B}, while the archived Llama pair uses \texttt{Pair A}. Run identifiers such as \texttt{M0-B\_installation} and \texttt{M+-A\_reversal} should be read through this mapping.

\paragraph{Checkpoints.} Llama pair comprises matched Llama-3-8B OpenHermes checkpoints before and after DPO on HH-RLHF \texttt{harmless-base}. Mistral pair comprises OpenHermes-2.5-Mistral-7B and the same checkpoint with its harmlessness-DPO adapter merged. Both have 32 transformer layers and residual dimension 4{,}096. $M_{\mathrm{rev}}$ is an auxiliary Pair-A checkpoint trained with preferred and rejected responses exchanged. All endpoints and tokenizers are pinned by revision, hash-identified, and validated before inference. Full checkpoint identities, adapter revisions, and training hyperparameters are recorded in the release manifest.

\paragraph{Partitions.} After removing 22 duplicate pairs, HarmfulQA yields 1{,}938 unique prompts, deterministically partitioned into 1{,}378 for direction construction, 200 for operating-window calibration, 300 for held-out evaluation, and 60 for pipeline development. Partitions are disjoint by source ID and by normalized prompt hash. The held-out partition is inaccessible during construction, calibration, coefficient selection, rubric development, and debugging. Transfer panels are 200 frozen HH-RLHF \texttt{harmless-base} test prompts and 200 deterministically selected AdvBench requests; neither influences any protocol choice.

\paragraph{Generation.} All conditions use endpoint-bound chat templates and greedy decoding, with maximum input and output lengths of 2{,}048 and 768 tokens. Generation stops on endpoint-specific EOS or EOT tokens or on a leaked subsequent instruction header, and all stopping metadata are recorded so that maximum-length termination can be separated from natural termination.

\paragraph{Evaluation.} A frozen Qwen3.5-27B evaluator, pinned by revision and served with vLLM in bfloat16, emits structured integer scores in $\{0,\ldots,10\}$ with justifications under fixed per-metric rubrics. Direct emitted scores are used; probability-weighted scores are never used. Deterministic request seeds, a fixed maximum of two retries, and explicit abstention on repeated schema failure are fixed in the protocol manifest. Abstentions are recorded and never imputed.

\paragraph{Validity screen.} A response is invalid if it is empty or abnormally short, leaks role or template tokens, or exhibits severe character-, punctuation-, subword-, or line-level repetition. The screen is deterministic and identical across all conditions. Invalid responses are excluded from behavioral means but retained in integrity-rate denominators; integrity rates therefore remain visible, while behavioral estimates remain conditional on valid, scored outputs.

\paragraph{Estimation.} Every contrast is computed over source IDs that are observed, valid, and scored in every condition the contrast requires. Intervention effects pair an intervention with its own unsteered baseline; gap recovery additionally requires both endpoints; specificity additionally requires all five random arms. These intersections are nested and are reported separately rather than pooled, which is why the same nominal quantity can carry slightly different values and sample sizes in different tables. Confidence intervals are two-sided 95\% percentile intervals from 10{,}000 paired prompt-bootstrap replicates resampling source IDs. Datasets are never pooled.

\paragraph{Data and code availability.} All endpoints, datasets, activation artifacts, evaluator protocols, code states, configurations, environments, and analysis inputs are revision- or hash-bound and archived with SHA-256 inventories. Every displayed value is traceable to an included, hash-verified analysis artifact. Tables whose per-record inputs are included can additionally be recomputed without re-running generation or scoring; larger omitted inputs are identified by immutable hashes and provenance records. Public release details will be provided separately.

\section{Additional Methodological Clarifications}

The following clarifications summarize design choices, interpretation, and limits, with references to the supporting analyses.

\paragraph{Q1. What does ``two-way control'' mean, and is it a single global direction?}
D-STEER adds a DPO-associated direction to the instruction-tuned endpoint to install safer behavior and subtracts it from the DPO endpoint to remove part of that behavior. At each selected layer, the direction is the mean paired activation difference between the two checkpoints on identical construction prompts. The primary intervention retains a separate vector at each of five layers; it does not collapse them into one global vector or exactly invert the DPO weight update. Held-out effects and specificity are reported in Table~\ref{tab:supp_specificity} and Figure~\ref{fig:supp_heldout_specificity}.

\paragraph{Q2. Why construct directions at the final prompt token?}
This is the boundary between prompt prefill and response generation: both checkpoints have consumed the same token sequence, but neither has generated an answer. Differencing their states controls prompt context and token identity and captures checkpoint-dependent processing of the same request. It does not isolate intent or a safety-only representation.

\paragraph{Q3. Why intervene at the final five layers?}
Late layers show elevated checkpoint displacement and substantial directional coherence (Figure~\ref{fig:supp_geometry}; Table~\ref{tab:supp_conversion}). The independent depth sweep also finds that intended movement grows as the contiguous late-layer block widens before flattening (Table~\ref{tab:supp_depth}). Because total injected activation grows with depth, this supports the practical efficacy of multilayer steering, not a dose-matched causal ranking of layers or a universal choice of five layers.

\paragraph{Q4. Why does the primary intervention act on both prompt and generated tokens?}
The primary protocol tests sustained actuation across the response, with coefficients calibrated for that full token scope. Scope changes cumulative intervention dose: the independent fixed-coefficient sweep shows that extending actuation can strengthen behavioral movement but also increase invalidity and quality loss (Table~\ref{tab:supp_breadth}). That sweep motivates scope-specific calibration; it does not isolate scope at the primary protocol's own coefficients.

\paragraph{Q5. How do relative normalization and raw-displacement scaling differ?}
Relative normalization expresses each layer's perturbation as a fraction of its typical activation magnitude; raw scaling retains the mean construction-set displacement, so a coefficient of one injects one such displacement per layer. They preserve each layer's direction but change its gain. Table~\ref{tab:supp_conversion} gives the within-protocol conversion; because the conversion varies by layer, one scalar cannot generally match the entire multilayer intervention. The independent raw implementation also changes construction data, depth, scope, evaluator, and usability criteria. Its twelve Mistral effects support directional robustness (Table~\ref{tab:supp_raw_grid}), not a controlled comparison of normalization schemes.

\paragraph{Q6. How were coefficients selected, and why need installation and removal not be symmetric?}
Selection uses only the frozen 200-prompt calibration split. Each direction's usable window is the contiguous-from-zero region meeting the prespecified integrity and quality limits; within it, selection maximizes intended harmfulness movement. Different starting checkpoints can have different usability frontiers, so installation and removal are calibrated separately. All four selected coefficients lie at their usable frontiers and are therefore constraint-limited, not unconstrained optima. The 300-prompt held-out panel remains sealed until these choices are frozen (Figure~\ref{fig:supp_operating_window}; Table~\ref{tab:supp_window_sensitivity}).

\paragraph{Q7. What is an invalid generation, and does filtering remove selection bias?}
The deterministic screen rejects empty or abnormally short outputs, leaked role or template tokens, and severe repetition. Invalid outputs leave behavioral means but remain in integrity-rate denominators; maximum-token termination is tracked separately (Table~\ref{tab:supp_conditions}). These reports expose generation failures but do not eliminate selection bias: behavioral estimates remain conditional on the prompts yielding valid, scored outputs in every required arm, and failures need not be random.

\paragraph{Q8. Is increased refusal itself treated as a safety improvement?}
No. Direct harmful enablement is the primary behavioral outcome; refusal is supporting evidence. Quality, invalidity, and maximum-token termination are assessed separately, so refusal alone is not the success criterion. These measurements do not establish harmlessness across all contexts or rule out over-refusal on benign prompts.

\paragraph{Q9. What do the five random directions establish?}
They test whether equal-size generic perturbations reproduce the real direction's effect. For each fixed seed, independent isotropic Gaussian directions are sampled at each steered layer and rescaled to the real direction's per-layer norm, with the coefficient, layers, scope, prompts, and decoding held fixed. Five seeds are a pragmatic, predeclared compute-versus-coverage choice, not a generally sufficient or optimal number. Seed-level effects and their spread are reported separately; specificity intervals bootstrap paired prompts rather than treating seeds as a large inferential sample (Tables~\ref{tab:supp_specificity}--\ref{tab:supp_seeds}; Figure~\ref{fig:supp_heldout_specificity}).

\paragraph{Q10. Does the direction generalize beyond HarmfulQA?}
Directionally, harmfulness and refusal move as intended in all eight HH-RLHF and AdvBench contrasts, and every harmfulness interval excludes zero (Table~\ref{tab:supp_transfer}; Figure~\ref{fig:supp_transfer}). However, fifteen of eighteen constituent transfer conditions exceed at least one HarmfulQA-calibrated integrity threshold (Table~\ref{tab:supp_gates}; Figure~\ref{fig:supp_usability}). Those estimates describe the valid, scored intersections, not universally usable deployment settings. Directional transfer is stronger than transfer of the original operating window.

\paragraph{Q11. Why do Mistral and Llama differ?}
The experiments establish checkpoint and implementation dependence, not its full mechanistic cause. The primary calibrated protocol establishes both directions with specificity for both pairs. The independent raw stress test is stronger and more stable for Mistral; Llama joint intersections can fall to 28 prompts, and only two of six specificity intervals exclude zero (Table~\ref{tab:supp_raw_spec}). The raw audit's controls also use last-five-layer rather than its primary last-seven-layer actuation. These limits preclude a model-independent effect-size or full-replication claim.

\paragraph{Q12. Does removal reproduce reversed-preference training?}
Steering-based removal and reversed-preference training agree in direction across four panels, supporting access to preference-relevant behavior. Their magnitude ratio varies by panel, and one intervention changes activations while the other changes weights. This is directional corroboration, not equivalence between steering and retraining (Figure~\ref{fig:supp_reverse_training}; Table~\ref{tab:supp_reverse}).

\paragraph{Q13. Why can related estimates have different sample sizes or values?}
Each contrast uses the strict intersection of source IDs observed, valid, and scored in all required arms. Intervention effects require the steered arm and its baseline; gap recovery additionally requires both endpoints; specificity additionally requires all five random arms. These nested intersections ensure within-contrast pairing but can change the retained prompts, sample size, and estimate across contrasts.

\paragraph{Q14. Why is gap recovery descriptive rather than the selection target?}
Gap recovery divides the intervention effect by the unsteered endpoint gap, helping interpret its scale. A small or panel-dependent denominator can make this ratio unstable. Coefficients are therefore selected by direct harmfulness movement within the usable window, not by maximizing recovery.

\paragraph{Q15. Does a frozen evaluator remove evaluator bias?}
No. Freezing the evaluator revision, rubrics, decoding, seeds, retry policy, and abstention rules standardizes scoring within comparisons; it does not establish unbiased or expert-equivalent judgments. Direct emitted scores are used, and abstentions are never imputed. The raw audit additionally finds nontrivial metric changes when identical generations are rescored. Prompt-bootstrap intervals do not capture all evaluator, checkpoint-training, or direction-construction uncertainty.

\paragraph{Q16. What kinds of data overlap are controlled?}
HarmfulQA is deduplicated and split by source ID and normalized prompt hash into construction, calibration, held-out, and pipeline-development partitions. The held-out split is inaccessible during construction, coefficient selection, rubric development, and debugging; transfer panels do not influence primary choices. These controls address within-study reuse, not possible overlap with checkpoint pretraining or earlier post-training data.

\paragraph{Q17. What can be reproduced from the archived artifacts?}
Checkpoints, datasets, activations, evaluator protocols, code states, configurations, environments, and analysis inputs are revision- or hash-bound. Included per-record artifacts permit recomputation of their corresponding analyses; larger omitted inputs are identified by immutable hashes and provenance records. This distinguishes an archived analysis trail from a fully self-contained reproduction of generation and scoring. Public release details will be provided separately.

\paragraph{Q18. What is causal, and what does the evidence not establish?}
Adding or subtracting a frozen direction while holding weights, prompts, and decoding fixed is a direct activation intervention. Paired behavioral estimates are conditional on the valid, scored intersection. Norm-matched controls support directional specificity for the tested settings, and primary integrity and quality gates constrain degradation; neither eliminates every alternative explanation or rescues transfer conditions that fail those gates. Geometry and cross-protocol comparisons remain descriptive. The results do not establish a universal coefficient, exact reversal of DPO, a single safety direction, or a complete mechanism of alignment; they show bidirectional accessibility of a meaningful component of DPO-associated safety behavior under the studied white-box conditions.

\end{document}